\documentclass[journal]{IEEEtran}
\usepackage{amssymb}

\usepackage[tight,footnotesize]{subfigure}  
\usepackage{amsmath}
\usepackage{amsfonts}
\usepackage{amssymb}

\usepackage{mathrsfs,amsfonts,dsfont}
\usepackage{bm}

\ifCLASSINFOpdf
   \usepackage[pdftex]{graphicx}
\else
   \usepackage[dvips]{graphicx}
\fi

\begin{document}
%
\title{Collaborative Representation Classification Ensemble for Face Recognition}
%
%
%

\author{Xiao Chao~Qu, Suah Kim, Run Cui and Hyoung Joong~Kim \thanks{Xiao Chao Qu is with the Center for Information Security Technologies (CIST), Korea University, Seoul 136171, Korea (e-mail: quxiaochao@gmail.com).}
  \thanks{Suah Kim is with the Center for Information Security Technologies (CIST), Korea University, Seoul 136171, Korea (e-mail: suahn@gmail.com).}
   \thanks{Run Cui is with the Center for Information Security Technologies (CIST), Korea University, Seoul 136171, Korea (e-mail: cuirun@korea.ac.kr).}
 \thanks{Hyoung Joong Kim is with the Center for Information Security Technologies (CIST), Korea University, Seoul 136171, Korea (e-mail: khj-@korea.ac.kr).}}
\maketitle

\begin{abstract}
Collaborative Representation Classification (CRC) for face recognition attracts a lot attention recently due to its good recognition performance and fast speed. Compared to Sparse Representation Classification (SRC), CRC achieves a comparable recognition performance with 10-1000 times faster speed. In this paper, we propose to ensemble several CRC models to promote the recognition rate, where each CRC model uses different and divergent randomly generated biologically-inspired features as the face representation. The proposed ensemble algorithm calculates an ensemble weight for each CRC model that guided by the underlying classification rule of CRC. The obtained weights reflect the confidences of those CRC models where the more confident CRC models have larger weights.  The  proposed weighted ensemble method proves to be very effective and improves the performance of each CRC model significantly. Extensive experiments are conducted to show the superior performance of the proposed method.
\end{abstract}


%
\IEEEpeerreviewmaketitle

\section{Introduction}
\label{sec:intro}

Face recognition is one of the hottest research topics in computer vision due to its wide range of applications, from public security to personal consumer electronics. Although signicicant improvement has been achieved in the past decades, a reliable face recognition system for real life environments is still very challenging to build due to the large intra-class facial variations, such as expression, illumination, pose, aging and the small inter-class facial differences~\cite{zhenhua}.

For a face recognition system, face representation and classifier construction are the two key factors. face representation can be divided into two categories: holistic feature based and local feature based. Principle Component Analysis (PCA) based Eigenface~\cite{turk1991face} and Linear Discriminative Analysis (LDA) based Fisherface~\cite{belhumeur1997eigenfaces} are the two most famous holistic face representations. PCA projects the face image into a subspace such that the most variations are kept, which is optimal in terms of face reconstruction. LDA considers the label information of the training data and linearly projects face image into a subspace such that the ratio of the between-class scatter over the within-class scatter is maximized. Both PCA and LDA projects the face image into a low dimensional subspace on which the classification is easier. It is based on an assumption that the high dimensional face images lie on a low dimensional subspace or sub-manifold. Therefore, it is beneficial to first project the high dimensional face image into that low dimensional subspace to extract the main structure of the face data and reduce the impact of the unimportant factors, such as illumination changes. Many other holistic face representations have been proposed later, including Locality Preserving Projection (LPP)~\cite{niyogi2004locality}, Independent Component Analysis (ICA)~\cite{bartlett2002face}, Local Discriminant Embedding (LDE)~\cite{chen2005local}, Neighborhood Preserving Embedding (NPE)~\cite{he2005neighborhood}, Maximum margin criterion (MMC)~\cite{li2006efficient} and so on.

The holistic face representation is known to be sensitive to expression, illumination, occlusion, noise and other local distortions. The local face representation which extracts features by using local information is shown to be more robust against those factors. The most commonly used local features in face recognition include Local Binary Pattern (LBP)~\cite{ahonen2006face}, Gabor Wavelets~\cite{liu2002gabor}, Scale-Invariant Feature Transform (SIFT)~\cite{lowe2004distinctive}, Histogram of Oriented Gradients (HOG)~\cite{dalal2005histograms} and so on.

To classify the extracted representations of faces into correct classes, a classier needs to be constructed. Many classifiers have been proposed and the most widely used classifier is the Nearest neighbor classifier (NN) and it is improved by Nearest Feature Line (NFL)~\cite{li1999face}, Nearest Feature Plane (NFP)~\cite{chien2002discriminant} and Nearest Feature Space (NFS)~\cite{chien2002discriminant} in different ways. Recently, Sparse Representation Classification (SRC)~\cite{wright2009robust} is proposed and shows good recognition performance and  is robust to random pixel noise and occlusion. SRC codes the test sample as a sparse linear combination of all training samples by exposing an $l_1$-norm constraint on the resulting coding coefficients. The $l_1$-norm constraint is very expensive which is the main obstacle of applying SRC in large scale face recognition systems. Lately, Collaborative Representation Classification (CRC)~\cite{zhang2011sparse} is proposed which achieves comparable performance to SRC and has a much faster recognition speed. The author in~\cite{zhang2011sparse} finds that it is the collaborative representation not the $l_1$-norm constraint that is important in the classification process. By replacing the slow $l_1$-norm with a much fast $l_2$-norm constraint, CRC codes each test sample as  a linear combination of all the training faces with a closed-form solution. As a result, CRC can recognize a test sample 10-1000 times faster than SRC as shown in~\cite{zhang2011sparse}.

In this paper, we propose to ensemble several CRCs to boost the performance of  CRC. Each CRC is a weak classifier are combined to construct the strong classifier named ensemble-CRC. For each test sample, several different face representations are extracted. Then, severl CRCs are used to make the classification using those face representations. A weight is then calculated and assigned to each CRC by considering the reconstruction residue characteristics. By analyzing the magnitude relationship between reconstruction residues of different classes, the highly correct CRC can be identified.  Large weights are assigned to those highly correct CRCs and small weights are assigned to the rest CRCs. Finally, the classification is obtained by a weighted combination of the reconstruction residues of all CRCs.

One key factor to the success of  ensemble learning is the significant diversity among the weak classifiers. For example, if different CRC makes different errors for test samples, then, the combination of many CRCs tends to yield much better results than each CRC. To this end, some randomly generated biologically-inspired face representation will be used. Biologically-inspired features have generated very competitive results in a variety of different object and face recognition contexts~\cite{lecun1998gradient},~\cite{serre2005object},~\cite{cox2011beyond}. Most of them try to build artificial visual systems that mimic the computational architecture of the brain. We use the similar model as in~\cite{jarrett2009best}, in which the author showed that the randomly generated biologically-inspired features perform surprisingly well, provided that the proper non-linearities and pooling layers are used. The randomly generated biologically-inspired model is shown to be inherently frequency selective and translation invariant under certain convolutional pooling architectures~\cite{saxe2011random}. It is expected that different randomly generated biologically-inspired features may generate different face representations (e.g., corresponds to different frequencies). Therefore, the proposed ensemble-CRC can obtain the significant diversity which is highly desired.

The rest of the paper is organized as follows. Section~\ref{Proposed} introduces the proposed ensemble-CRC method.  Section~\ref{Experiment} conducts extensive experiments to verify the effectiveness of ensemble-CRC. Finally, Section~\ref{Conclusion} concludes the paper.

\section{Proposed Method}
\label{Proposed}
\subsection{Ensemble-CRC}
First, we briefly introduce CRC. CRC codes a test sample using all the training samples linearly and pose an $l_2$ constraints on the coding coefficients. Then, the reconstruction of the test sample is formed by linearly combine the training samples from a specific class utilizing the corresponding coding coefficients. The test sample is classified into the class that has the smallest reconstruction error.

More specifically, suppose there are $n$ training samples from $c$ different classes.  For each class $j = 1, 2,... c$, there are $n_j$ training samples. The $i$th training sample of class $j$ is denoted as $x_{ji} \in \mathds{R}^{m}$ where $m$ is the feature's dimensionality. Let $\bm{A} = [ \bm{A}_1, \bm{A}_2, ..., \bm{A}_c] \in \mathds{R}^{m \times n}$ be the set of entire training samples, where $\bm{A_j} = [x_{j1}, x_{j2}, ..., x_{jn_j}] \in \mathds{R}^{m \times n_j}$ is composed of training samples from class $j$. For a given test sample $y$, CRC solves the following problem

\begin{equation}
\hat{\alpha} = \text{arg}\  \text{min} \{ || y - A\alpha ||_2^2 + \lambda ||\alpha||_2^2 \},
\end{equation}
where $\lambda$ is the regularization parameter. The solution of the above problem can be obtained analytically as

\begin{equation}
\label{alpha}
\hat{\alpha} = (A^TA+\lambda I)^{-1}A^Ty.
\end{equation}

Let $\bm{P} = (A^TA+\lambda I)^{-1}A^T$. It can be seen that $\bm{P}$ is independent of the test sample $y$ and can be pre-calculated. For each test sample, we only need simply project $y$ onto $\bm{P}$ to obtain the coding coefficients. To make the classification of $y$, the reconstruction of $y$ by each class should be calculated. For each class $j$, let $\delta_j: \mathds{R}^{n} \rightarrow \mathds{R}^{n}$ be the characteristic function that keeps the coefficients of class $j$ and assigns the coefficients associated with other class to be $0$. The reconstruction of $y$ by the class $j$ is obtained as $\hat{y}_j = A\delta_j(\hat{\alpha})$. The reconstruction error of class $j$ is obtained by

\begin{equation}
\label{re}
e_j = ||y - \hat{y}_j||_2^2 = || y - A\delta_j(\hat{\alpha}) ||_2^2
\end{equation}

CRC classifies $y$ into the class that has minimum reconstruction error.

The proposed ensemble CRC utilizes multiple CRCs and combines them together to obtain a final classification. Assume there are $k$ different face representations extracted from each face, and $k$ training set can be formed as $\bm{A}^1,..., \bm{A}^k$ and $A^k = [ \bm{A}_1^k, \bm{A}_2^k, ..., \bm{A}_c^k] \in \mathds{R}^{m \times n}$. Then, $k$ projection matrix $\bm{P}^1,..., \bm{P}^k$ can be obtained using $\bm{A}^1,..., \bm{A}^k$. For a test sample $y$, $k$ different representations are extracted and denoted as $y^1,..., y^k$. For each set of $(y^k,\bm{P}^k,\bm{A}^k)$, the coding coefficients $\alpha^k$ can be obtained using Equation~(\ref{alpha}) and the corresponding reconstruction errors $e_j^k$ can be obtained using Equation~(\ref{re}).

Different face representation has different performance for a particular test sample, therefore,  proper weights should be assigned to different CRCs given the test sample. Notice that CRC determines the class of the test sample by selecting the minimum classification error. If the correct class produces small reconstruction error and all other incorrect classes produce large reconstruction errors, CRC makes correct classification easily in this situation. However, when some incorrect classes  produce similar or smaller reconstruction error compared with the correct class, CRC may make wrong classification in this situation. In the latter situation, the reconstruction error of the correct CRC is usually among the several small reconstruction errors. In summary, CRC has high fidelity of correct classification when there is only one small reconstruction error and CRC has low fidelity of correct classification when there are several small reconstruction errors.  We utilize this observation to guide the calculation of the weights. For each representation, the smallest (denoted as $e_s$) and the second smallest (denoted as $e_{ss}$) reconstruction errors are picked, then the difference value between the two reconstruction errors is calculated as $d = e_{ss} - e_s$. Each representation has its difference value and $k$ difference values can be obtained as $d^1,..., d^k$. Then, the weight for the $k$th CRC can be calculated as

\begin{equation}
w^k = \frac{d^k}{d^1 + d^2 + ... + d^k}.
\end{equation}

It is obvious that the larger the difference, the larger the weight. After obtaining all the weight, the reconstruction error of class $j$ is calculated as

\begin{equation}
e_j = w^1*e_j^1 + w^2*e_j^2 + ... + w^k*e_j^k.
\end{equation}

The ensemble-CRC will assign the test sample into the class where the combined reconstruction error has minimum value.

\subsection{Randomly Generated Biologically-Inspired Feature}
The  biologically-inspired features used in the proposed ensemble-CRC are similar in form as the  biologically-inspired features in~\cite{jarrett2009best}. The feature extraction process includes four layers: filter bank layer, rectification layer, local contrast normalization layer and pooling layer. Different Biologically-inspired features can be obtained by modifying the structure of the extraction process or using different model parameters.  The details of each layer are introduced in the following.

\begin{itemize}
  \item Filter bank layer. The input image is convolved with a certain number of filters. Assume the input image $x$ has size $n_1 \times n_2$ and each filter $k$ has size $l_1 \times l_2$, the convolved output (or feature map) $y$ will have size $n_1-l_1+1 \times n_2-l_2+1$. The output can be computed as
       \begin{equation}
        y = g \times tanh(k \otimes x)
       \end{equation}
       where $\otimes$ is the convolve operation, tanh is the hyperbolic tangent non-linearity function and $g$ is a gain factor.
  \item Rectification layer. This layer simply applies the absolute function to the output of the filter bank layer as $y = |y|$.
  \item Local contrast normalization layer.  Local subtractive and divisive normalization are performed which enforces the local competition between adjacent features in a feature map. More details can be found in~\cite{pinto2008real}.
  \item Pooling layer. The pooling layer transforms the joint feature representation into a more robust feature which  achieves invariance to transformations,  clutter and small distortions. Max pooling and average pooling can be used. For max pooling, the max value of a small non-overlapping region in the feature map is selected. All other features in this small local region are discarded. The average pooling returns the average value of the small local region in the feature map. After pooling, the number of feature in feature maps are reduced. The reduction ratio is determined by the size of the local region.
\end{itemize}

it is shown in~\cite{jarrett2009best}  that the filters in the filter bank layer can be assigned with small random values and the obtained randomly generated features still achieve very good recognition performance in several image classification benchmark data sets.

The reason that we select the randomly generated biologically-inspired features in the proposed ensemble-CRC is twofold. First, it performs well in many different visual recognition problems, and second, the randomness in it provides some diverseness. It is shown that a necessary and sufficient condition for an ensemble of classifier to be more accurate than any of its individual members is if the classifiers are accurate and diverse~\cite{dietterich2000ensemble}.
\subsection{The Complete Recognition Process}
\begin{figure*}[tbp]
\centering
\includegraphics[width=7in]{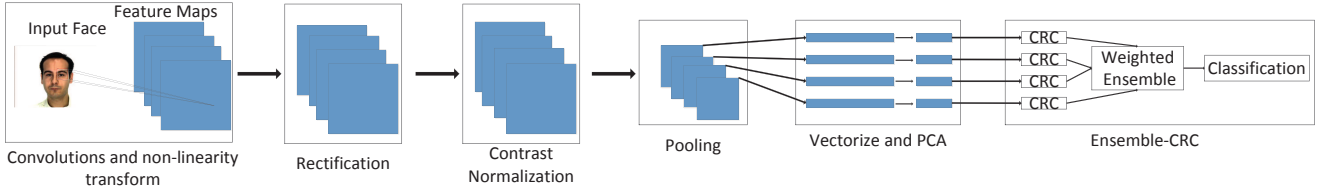}
\caption{The flowchart of the recognition process of the proposed ensemble-CRC. }
\label{flowChart}
\end{figure*}

The complete recognition process for a test face image is shown in Fig.~\ref{flowChart}. The input face image is first convolved with $k$ filters and then transformed non-linearly. As a result, $k$ feature maps are obtained, which are then rectified and normalized. Then, pooling is used to extract the salient features and reduce the feature map's size. Because the extract feature maps still have big size, we transform the $2$-D feature maps into $1$-D vectors and use PCA to reduce the dimensionality. After PCA, $k$ feature maps are transformed into $k$ face representations with reduced dimensionality. Up to now, we finish the extraction of different features. Next, the $k$ extracted features are used by $k$ CRCs, then, $k$ classification results are weighted combined to form the final classification result.

\section{Experiment}
\label{Experiment}
We compare the proposed ensemble-CRC with CRC~\cite{zhang2011sparse}, AW-CRC (Adaptive and Weighted Collaborative Representation Classification)~\cite{timofte2014adaptive}, SRC~\cite{wright2009robust}, WSRC (Weighted Sparse Representation Classification)~\cite{lu2013face} and RPPFE (Random Projection based Partial Feature Extraction)~\cite{ma2015random}. using AR~\cite{martinez1998ar} and LFW~\cite{LFWTech} face databases.

\begin{figure}[tbp]
  \centering
     \subfigure[AR ]{
        \includegraphics[width = 1.5 in]{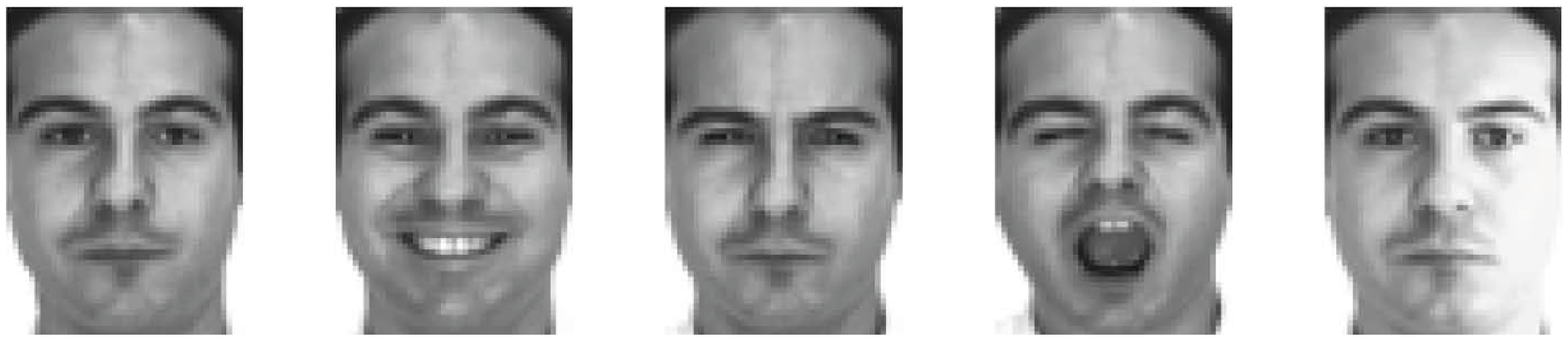}}
    \hspace{0.1in}
     \subfigure[ LFW ]{
        \includegraphics[width = 1.5 in]{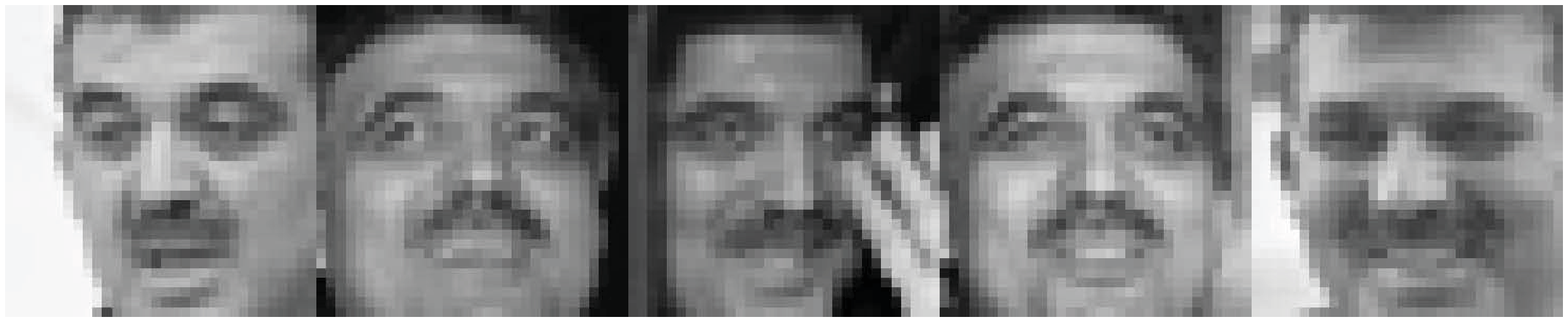}}
    \hspace{0.1in}
    \caption{The sample face images of  AR and LFW databases  }
    \label{samples}
\end{figure}

The AR database consists of over $4,000$ frontal face images from $126$ individuals. The images have different facial expressions, illumination conditions and occlusions. The images were taken in two separate sessions, separated by two weeks time. In our experiment, we choose a subset of the AR database consisting of $50$ male subjects and $50$ female subjects and crop image into the size of $64 \times 43$. For each subject, the seven images with only illumination change and expressions from Session one are used for training. The seven images with only illumination change and expressions from Session two are used for testing.

The Labeled Faces in the Wild (LFW) database is a very challenging database consists of faces with great variations in terms of lighting, pose, expression and age. It contains $13,223$ face images from $5,749$ persons. LFW-a is a subset of LFW that the face images are aligned using a commercial face alignment software. We adopt the same experiment setting in~\cite{zhu2012multi}. In detail, $158$ subjects in LFW-a that have no less than $10$ images are chosen. For each subject, $10$ images are selected in the experiment. Thus, there are in total $1,580$ images used in our experiment. Each image is first cropped to $121 \times 121$ and then resized to the size of $32 \times 32$. Five images are used for training and the other five images for testing.

In all the following experiment, the filter size used is $5 \times 5$, and all filters are randomly generated from a uniform distribution from $[-0.001, 0.001]$. The  non-linearity function used is $f(a) =1.7159 tanh(0.6667 a )$ as in~\cite{lecun1998gradient}. The pooling used is max pooling with size $2 \times 2$.

\subsection{Number of CRCs in Ensemble-CRC}
The number of weak classifiers in an ensemble classifier is very important to the performance of the ensemble classifier. The increase of the number of weak classifiers improve the performance of the ensemble classifier at first, but the performance of the ensemble classifier may degrade when too many weak classifiers are used. Also, the more the weak classier, the more the computation is needed. Next, we conduct several experiments on AR database to show the huge impact of the number of weak classifiers and try to find the best number experimentally.

\begin{figure}[tbp]
\centering
\includegraphics[width=3in]{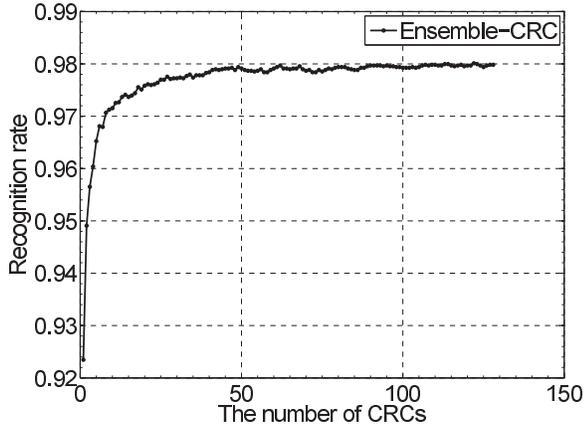}
\caption{The performance of ensemble-CRC with different number of CRCs. }
\label{NumCRC}
\end{figure}

We test the number of weak classifier from $1$ to $128$ and the dimension after PCA is set as $300$. We repeat the experiment $10$ times and the average result is reported in Fig.~\ref{NumCRC}. It can be seen that the recognition rate is $92.4\%$ when only one CRC is used. With eight CRCs included in ensemble-CRC, the performance increases rapidly to $97.1\%$. When $64$ CRCs are used in ensemble-CRC, the performance is around $98\%$, and more CRCs do not improve the performance further. We conclude that $64$ CRCs seem to be the best number of weak classifiers. All the rest experiments thus use $64$ CRCs in ensemble-CRC.

\subsection{Weighted VS. Non-Weighted Ensemble-CRC}
\begin{figure}[tbp]
    \centering
\includegraphics[width=3in]{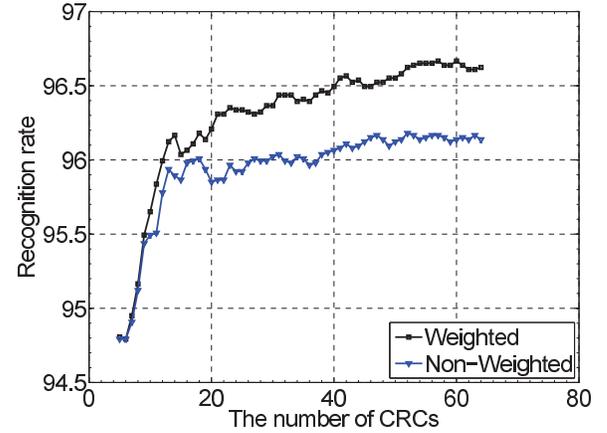}
    \caption{The performance comparison of the proposed weighted ensemble-CRC and the non-weighted ensemble-CRC }
    \label{comparisonWeight}
\end{figure}

In the proposed ensemble-CRC, a weight is calculated for each CRC. The weights can all be assigned to be $1$, and the obtained ensemble-CRC can be regarded as non-weighted ensemble-CRC. In the following, we compare the performance of the proposed weighted ensemble-CRC and the non-weighted ensemble-CRC on AR database, using the feature dimension of $100$. Fig.~\ref{comparisonWeight} shows that the weighted ensemble-CRC consistently outperforms the non-weighted ensemble-CRC.

\subsection{Performance Comparison With Other Methods}
\begin{figure}[tbp]
  \centering
     \subfigure[AR ]{
        \includegraphics[width = 3 in]{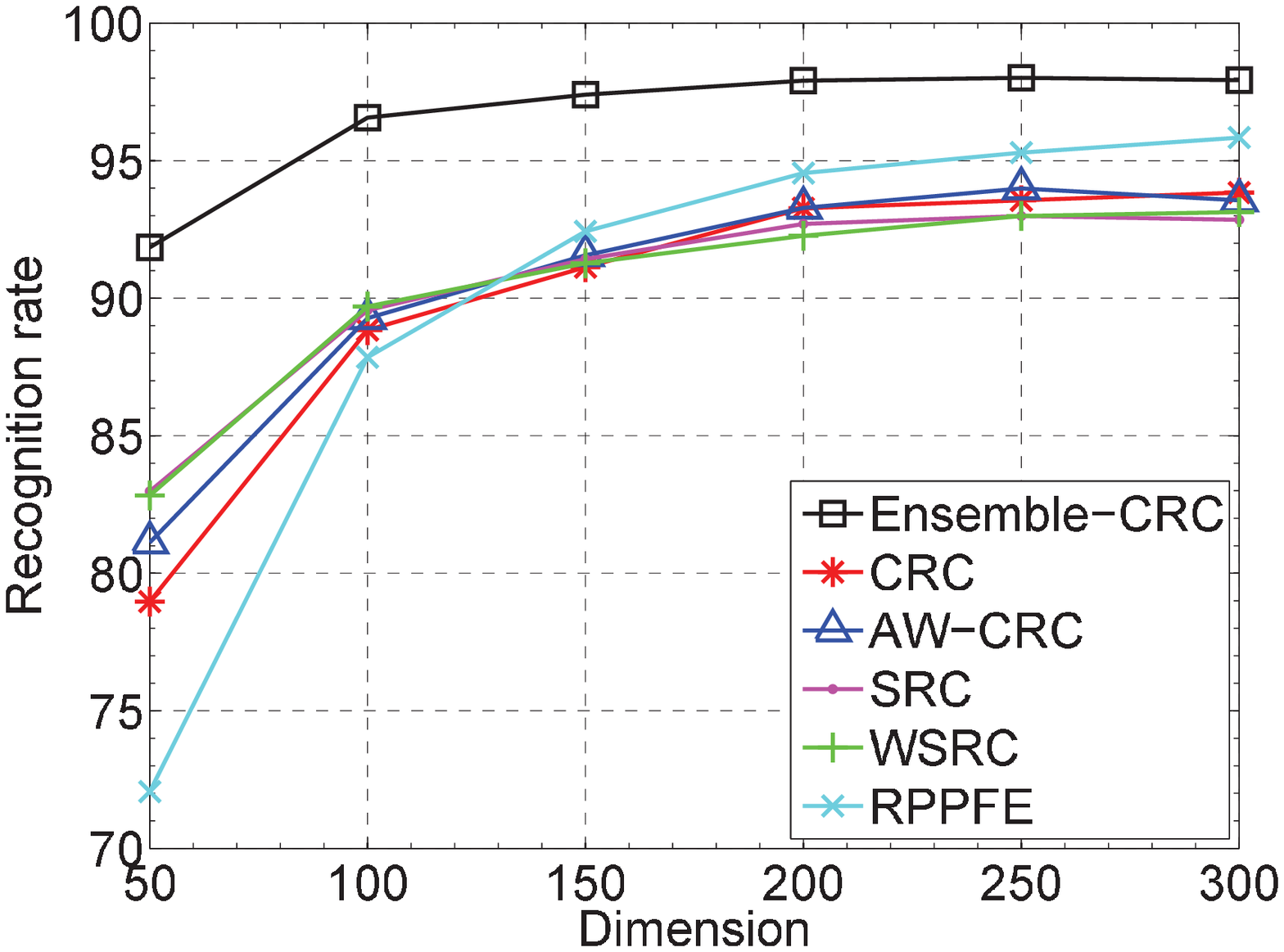}}
    \hspace{0.1in}
     \subfigure[LFW]{
        \includegraphics[width = 3 in]{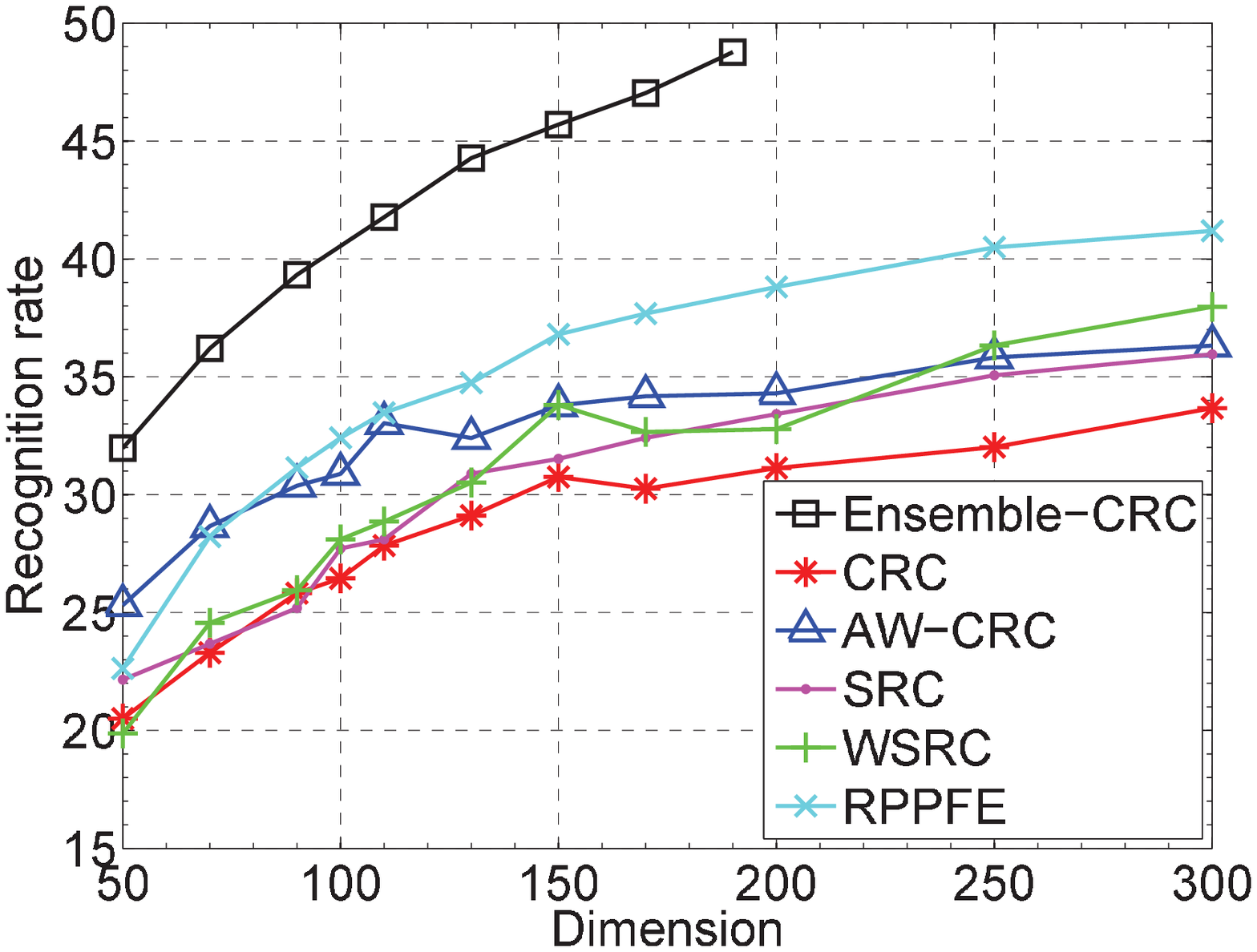}}
    \hspace{0.1in}
    \caption{The performance comparison of the proposed ensemble-CRC with CRC and AW-CRC on AR and LFW database.}
    \label{experimentResult}
\end{figure}

In the following, the proposed ensemble-CRC is compared with CRC, AW-CRC, SRC, WSRC  and RPPFE. Different feature dimensions are compared for each database as shown in Fig.~\ref{experimentResult}. For AR database, ensemble-CRC achieves the recognition rate of $91.85\%$ with feature dimension of $50$, which is $12.88\%$ higher than that of CRC ($78.97$), $10.73\%$ higher than that of AW-CRC ($81.1\%$), $8.87\%$ higher than that of SRC($82.98\%$), $9.02\%$ higher than that of WSRC($82.83\%$) and $19.79\%$ higher than that of RPPFE($72.06\%$). With the increase of the dimension, the performance of ensemble-CRC, CRC, AW-CRC, SRC, WSRC and RPPFE all increase gradually. The highest recognition rate of ensemble-CRC, CRC, AW-CRC, SRC, WSRC and RPPFE are $98.10\%$, $93.84\%$, $93.99\%$, $92.99\%$, $93.13\%$ and $95.84\%$ respectively. It is clear that the proposed ensemble-CRC outperforms all other methods.

The LFW database is quite difficult. The highest recognition rate obtained by CRC, AW-CRC, SRC, WSRC and RPPFE  is $33.67\%$, $36.32\%$, $35.95\%$ and$37.97\%$, which are much lower than that of AR database. The proposed ensemble-CRC achieves the highest recognition rate of $48.77\%$ which is much higher than that of CRC, AW-CRC, SRC, WSRC and RPPFE. Due to the pooling operation, the dimension for each randomly generated biologically-inspired feature is constrained to be $190$. However, the recognition rate may be higher if higher dimension of randomly generated biologically-inspired feature can be used (e.g., larger input image size), which can be inferred from the recognition rate curve of ensemble-CRC.

\section{Conclusion}
\label{Conclusion}
In this paper, a novel face recognition algorithm named ensemble-CRC is proposed. Ensemble-CRC utilizes the randomly generated biologically-inspired feature to create many high-performance and diverse CRCs which are combined using a weighted manner. The experimental result shows that the proposed ensemble-CRC outperforms the  CRC, AW-CRC, SRC, WSRC and RPPFE.

\bibliographystyle{IEEEtran}
\bibliography{IEEEabrv,my}




%








\end{document}